\documentclass[10pt,twocolumn,letterpaper]{article}

\usepackage{3dv}
\usepackage{times}
\usepackage{epsfig}
\usepackage{graphicx}
\usepackage{amsmath}
\usepackage{amssymb}

\usepackage{xcolor}
\usepackage[normalem]{ulem}
\usepackage{multirow}
\usepackage[inline]{enumitem}
\usepackage{authblk}

\usepackage[pagebackref=true,breaklinks=true,letterpaper=true,colorlinks,bookmarks=false]{hyperref}

\threedvfinalcopy 

\def\threedvPaperID{} 

\definecolor{tableRed}{RGB}{255,0,0}
\definecolor{tableBlu}{RGB}{0,86,230}
\definecolor{tableGre}{RGB}{51,153,40}

\newcommand{\best}[1]{\textcolor{tableRed}{\textbf{#1}}}
\newcommand{\second}[1]{\textcolor{tableGre}{\textbf{#1}}}
\newcommand{\third}[1]{\textcolor{tableBlu}{\textbf{#1}}}

\ifthreedvfinal\fi
\begin{document}

\title{Detail Preserving Depth Estimation from a Single Image Using\\ Attention Guided Networks\vspace{-2ex}}

\renewcommand\Authsep{\qquad}
\renewcommand\Authand{}
\renewcommand\Authands{\qquad}
\author[1]{Zhixiang Hao}
\author[3]{Yu Li}
\author[4,5]{Shaodi You}
\author[1,2,*]{Feng Lu}
\affil[1]{State key Laboratory of Virtual Reality Technology and Systems, School of Computer Science and Engineering, Beihang University, Beijing, China}
\affil[2]{Beijing Advanced Innovation Center for Big Data-Based Precision Medicine \authorcr Beihang University, Beijing China}
\affil[3]{Advanced Digital Sciences Center, Singapore\quad $^{\text{4}}$Data61-CSIRO\quad $^{\text{5}}$Australian National University\vspace{0.5em}}
\affil[ ]{\texttt{\normalsize{\{haozx, lufeng\}@buaa.edu.cn}\qquad yul@illinois.edu\qquad shaodi.you@data61.csiro.au}}

\maketitle

\begin{abstract}
    Convolutional Neural Networks have demonstrated superior performance on single image depth estimation in recent years. These works usually use stacked spatial pooling or strided convolution to get high-level information which are common practices in classification task. However, depth estimation is a dense prediction problem and low-resolution feature maps usually generate blurred depth map which is undesirable in application. In order to produce high quality depth map, say clean and accurate, we propose a network consists of a Dense Feature Extractor (DFE) and a Depth Map Generator (DMG). The DFE combines ResNet and dilated convolutions. It extracts multi-scale information from input image while keeping the feature maps dense. As for DMG, we use attention mechanism to fuse multi-scale features produced in DFE. Our Network is trained end-to-end and does not need any post-processing. Hence, it runs fast and can predict depth map in about 15 fps. Experiment results show that our method is competitive with the state-of-the-art in quantitative evaluation, but can preserve better structural details of the scene depth.
\end{abstract}

\let\thefootnote\relax\footnote{* Corresponding Author: Feng Lu}
\section{Introduction}

    \begin{figure}[t]
        \centering
        \includegraphics[width=\linewidth]{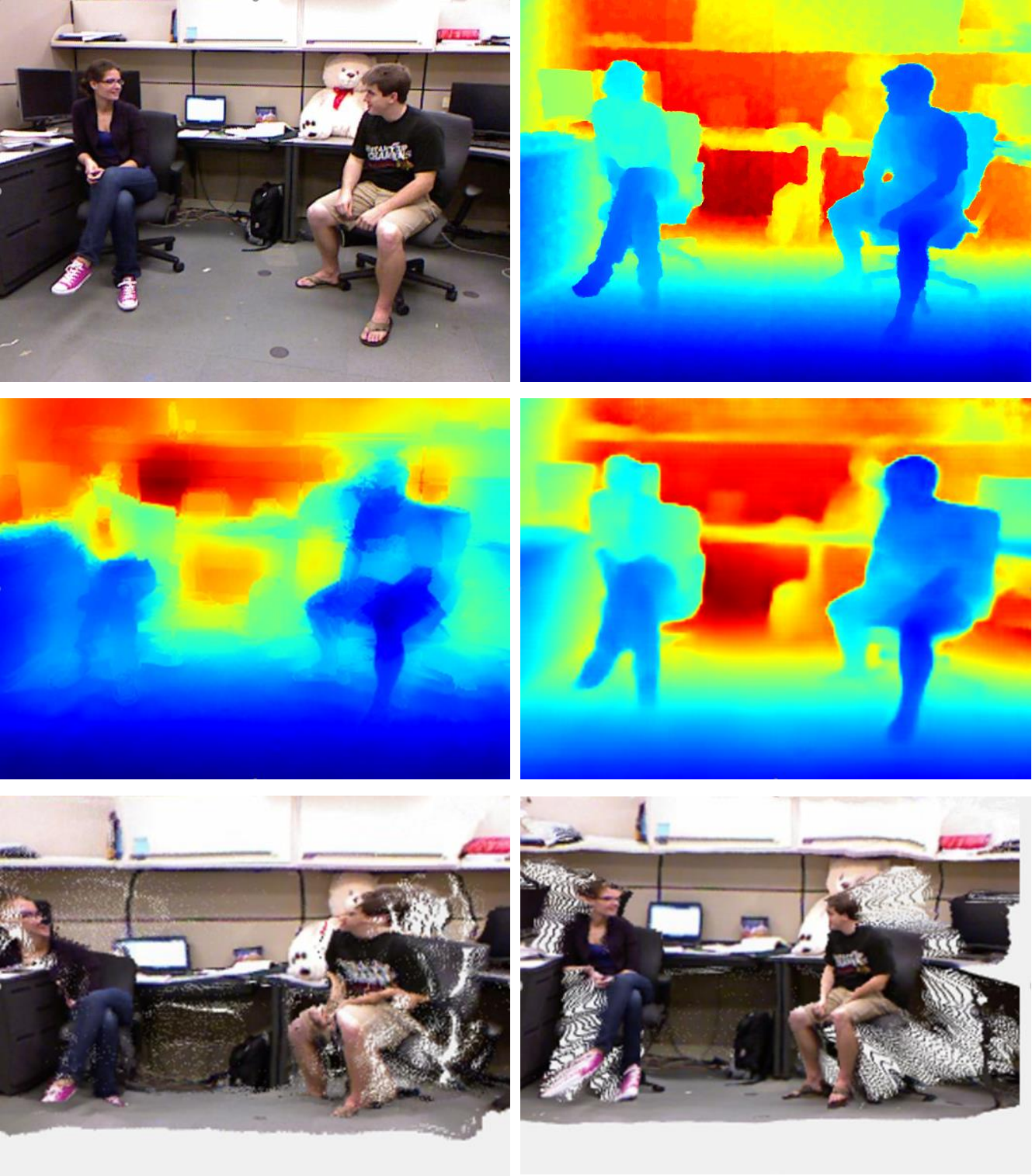}
        \caption{(\textbf{Top left}) input RGB image. (\textbf{Top right}) ground truth depth. (\textbf{Middle left}) depth estimated by Xu \etal~\cite{xu2017multi} (relative error: $0.092$). (\textbf{Middle right}) depth estimated by our method (relative error: $0.085$).
(\textbf{Bottom left}) point cloud from Xu \etal depth. (\textbf{Bottom right}) point cloud from ours depth. While the accuracy is comparable, our detail-preserving result performs better in depth related applications.}
        \label{fig:first}
        \vspace{-1.5em}
	\end{figure}
    
   	Single image depth estimation is an important task as it enables a variety of computer vision and graphics applications, such as 3D scene reconstructions~\cite{tateno2017cnn},    depth-aware image re-rendering~\cite{basha2013stereo}, and image refocus~\cite{anwardepth} \etc. It is given only a single RGB image as the input, however, aiming to estimate a depth map as output. In particular, indoor scenes have complex textures and structural variations which make it more difficult.
    
    Traditional methods use hand-crafted priors such as Karsch \etal~\cite{karsch2012depth} who use a depth transfer based on SIFT Flow~\cite{liu2011sift} and Liu \etal~\cite{liu2010single} who combined predicted semantic information with depth features. Recently, deep learning based methods have shown superior performance by directly learning from a large amount of data ~\cite{eigen2015predicting, eigen2014depth,liu2015deep,laina2016deeper,xu2017multi}.
    
    A high-quality depth map is expected to be both accurate and detailed. Existing deep neural network based methods have mostly focused on accuracy, but do not pay much attention to the important details~\cite{fu2018deep, kundu2018adadepth}. An example is shown in Fig.~\ref{fig:first}, while both Xu \etal and our method have comparable accuracy; only our method preserves the structural detail and produces reasonable results in 3D reconstruction. Other existing methods suffer from the same problem~\cite{laina2016deeper,eigen2014depth,chakrabarti2016depth}. The loss of the detail comes from the fact that CNN use an intensive number of strided convolutions and spatial poolings; it reduces the resolution and loses the accurate location information. Leading to a smoothed and low-resolution output. Long \etal~\cite{long2015fully} and  Ronneberger \etal~\cite{ronneberger2015u} have proposed methods to improve the resolution of feature maps. However, the detailed edges still fail to align with the image.
    
    To preserve structural details and produce a sharp output, we propose two network modules. The first is Dense Feature Extractor (DFE) which combines ResNet~\cite{he2016deep} with dilated convolution. DFE uses dilated convolution to extract multi-scale features from an image while keeping the feature maps dense. We also introduce a Depth Map Generator (DMG) module which uses attention mechanism to regress the feature maps to depth in which we can allocate compute resources toward the most informative parts of an input signal according to the context. By combining the proposed DFE and DMG, our network can extract multi-scaled dense while informative features and fuse them effectively. Quantitative experiments on NYU Depth V2 dataset~\cite{Silberman:ECCV12} show our method is competitive with the state-of-the-art. Moreover, projected to point cloud and bokeh generation show that our method can preserve better structural details of the scene depth.
    
    We summarize our contributions as follows:
    \begin{itemize}
    \item We propose a novel approach for predicting depth map from a single image which integrates a Dense Feature Extractor and attention mechanism.
    \item We propose a Fully Convolutional Network which can predict accurate depth map with errors competitive to the state-of-the-art on benchmark dataset, moreover depth map produced by proposed method preserves significantly more structural details that benefit various applications.
    \end{itemize}

\section{Related Work}

    Single image depth estimation is an important problem in computer vision and an active field of research. We focus on reviewing the monocular methods.

    Classic methods for depth estimation from a single image mainly relied on hand-crafted features or graphical models. Saxena \etal~\cite{saxena2006learning} use a multi-scale Markov Random Field (MRF) and linear regression to predict depth information from features extracted from a single image. Later, they extend their work into Make3D~\cite{saxena2009make3d}. Inspired by this work, Liu \etal~\cite{liu2010single} show that depth estimation can benefit from combining semantic information with depth features, where predicted labels are used as additional constraints to help the optimization task. Ladicky \etal~\cite{ladicky2014pulling} predict semantic labels and depth maps jointly in a classification approach. They train a classifier considering both the predicted semantic labels loss and depth loss. Karsch \etal~\cite{karsch2012depth} propose a depth transfer, a non-parametric method based on SIFT Flow~\cite{liu2011sift}, to reconstruct depth by transferring depth of image in the same dataset. Liu \etal~\cite{liu2014discrete} treat depth estimation as an optimization problem which is formulated as a Conditional Random Field (CRF) with continuous and discrete variable potentials.

    Recently, most of the state-of-the-art for many problems in computer vision is set by deep neural networks, especially CNNs~\cite{he2016deep,he2017mask, szegedy2016rethinking}. Not surprisingly, deep learning is highly effective for depth estimation. Eigen \etal~\cite{eigen2014depth} is the first to use CNNs for estimating depth from a single image. They propose a two-scale network in where one predicts coarse depth from the entire image, and the other refine the coarse prediction locally. This work is extended to predict normal and semantic labels using a three-scale network based on VGG~\cite{eigen2015predicting, simonyan2014very}. Liu \etal~\cite{liu2016learning} propose combining CNNs and CRF to learn the unary and pairwise potentials with CNNs. They predict depth at a superpixel level and train the CNNs with CRF loss, while Li \etal~\cite{li2015depth} and Wang~\cite{wang2015towards} use hierarchical CRFs to refine patch-wise CNNs prediction from superpixel down to pixel level. Roy~\cite{roy2016monocular} proposes the neural regression forest (NRF), which combines random forests and CNNs. Laina \etal~\cite{laina2016deeper} propose leveraging the power of pre-trained CNNs such as ResNet~\cite{he2016deep} to predict dense depth maps. Combined with fast up-projection, they archive state-of-the-art results on single image depth estimation.
    
    \begin{figure*}[th]
        \centering
        \includegraphics[width=\linewidth]{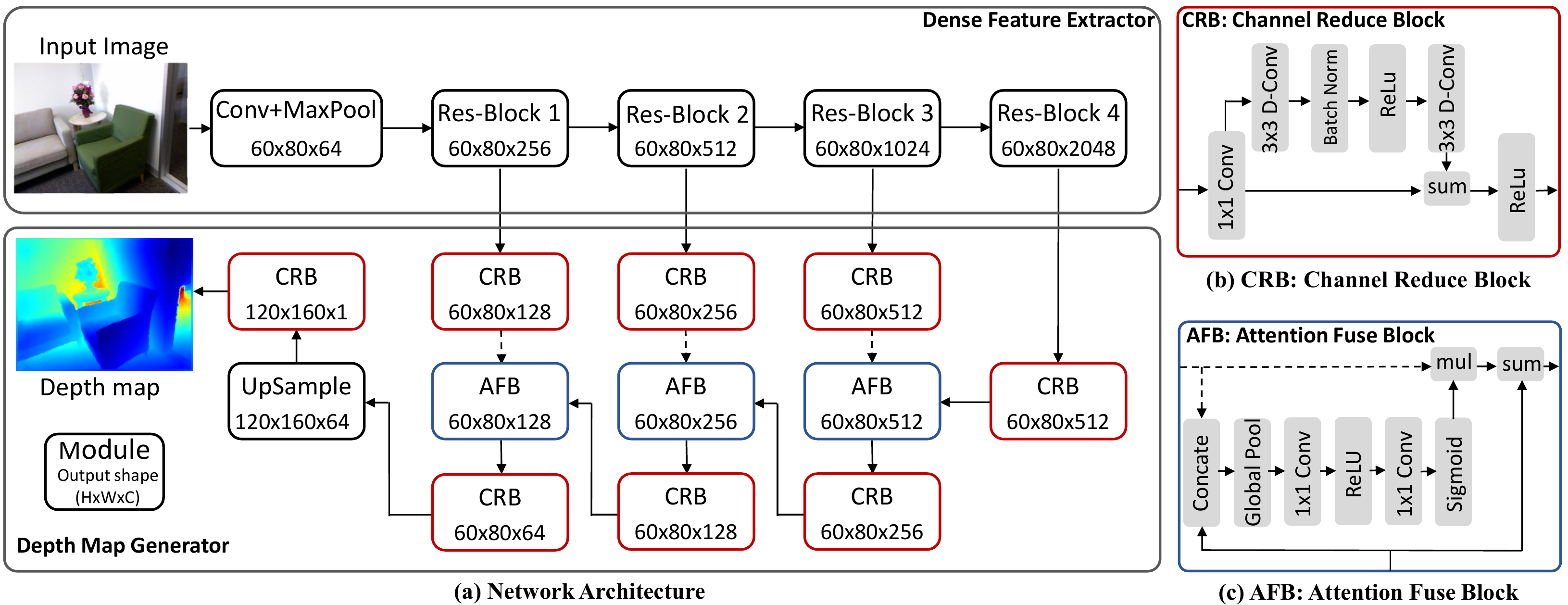}
        \caption{{\bf Network Architecture Overview.} (a) Network architecture. (b) Inner structure
         of Channel Reduce Block (CRB). (c) Inner structure of Attention Fuse Block (AFB). The dotted line and solid line distinguish
         the two different inputs in AFB, the D-Conv in CRB represents Dilated Convolution. The input image in the diagram has
         resolution of $240$x$320$.}
        \label{fig:network}
	\end{figure*}

    More recently, Li \etal~\cite{li2017two} use two-streamed CNNs that predicts both depth and depth gradients, and then fuse them together to produce an accurate depth map. Xu \etal~\cite{xu2017multi} fuse complementary information derived from multiple CNNs side outputs by using continuous CRFs. Later, they combine this work with attention model~\cite{xu2018structured}. Fu \etal~\cite{fu2018deep} introduce a spacing-increasing discretization (SID) strategy to discretize depth and translate depth estimation from a regression to a classification task. There are also some methods focus on unsupervised or semi-supervised learning by recovering a right view with a left view~\cite{garg2016unsupervised, kuznietsov2017semi}.

\section{Method}

    In this section, we first detailedly describe the proposed Dense Feature Extractor (DFE) and Depth Map Generator (DMG). Then, by combining the DFE and DMG, we introduce the complete architecture of our proposed network.
    
    \subsection{Dense Feature Extractor}
    
    Many previous depth estimation methods directly use networks designed for image classification tasks~\cite{laina2016deeper,eigen2015predicting,xu2017multi}, in which, networks output low-resolution feature maps. Hence, these methods suffer from low-resolution or over-smoothed output. To recover the high-resolution spatial detail from the low-resolution feature maps, previous methods \cite{laina2016deeper,eigen2014depth,xu2018structured} usually use the repeated combination of deconvolutional layers (or transposed convolution) which is time-consuming and complicate network training.
    
    Instead, we leverage the using of `dilated convolution' (or atrous convolution) to extract higher contextual information while keeping the spatial resolution of feature maps unchanged. Such technology is inspired by recent advance in semantic segmentation ~\cite{yu2015multi, chen2018deeplab}. Dilated convolution introduces a parameter called dilation rate. By adjusting dilation rate, it is convenient to control the resolution of feature maps and adjust the convolution kernel's field-of-view in order to extract multi-scale information. For instance, a higher dilation rate such $4$ or $8$ in our DFE performs convolution with a bigger filed-of-view which extracts higher-level information. In the case of two-dimensional signals such as image, for each location $i$ on the output $y$ and a filter $w$, dilated convolution is applied over the feature maps $x$ can be formalized as follows~\cite{chen2018encoder}:
    \begin{equation}
        y[i] = \sum_kx[i +r \cdot k] w[k] ,
    \end{equation}
    where $r$ is dilation rate. When $r=1$ dilated convolution is equal to standard convolution.

    In our proposed network, we combine dilated convolution and ResNet-101~\cite{he2016deep} to form DFE. As shown in Fig.~\ref{fig:network}, different from the original ResNet, we remove the global average pooling and fully connected layer at the end of ResNet and replace all standard convolution layers with dilated convolution layers in Res-Block 1 to Res-Block 4. Like the original ResNet, our DFE can be divided into five stages according to the size of filled-of-view. By introducing different dilation rate in different Res-Block, our DFE can extract multi-scale information as side outputs in each Res-Block. In the lower stage, \ie small dilation rate, the feature maps contain more low-level spatial information such as location and edge, however, it has poor semantic information due to its small field-of-view. While in the higher stage, \ie big dilation rate, it has bigger field-of-view, hence, more semantic information. Overall, our DFE produces multi-scale feature maps which can represent different information extracted from an input image.

    \subsection{Depth Map Generator}
    
   	To combine the multi-scale information produced by DFE, we introduce the DMG. DMG consists of Attention Fuse Block (AFB) and Channel Reduce Block (CRB). Fig.~\ref{fig:network} illustrate the structure of our DMG.
    
    {\bf Attention Fuse Block.} Attention mechanism allows us to allocate available compute resources towards the most important parts of an input signal according to the context~\cite{mnih2014recurrent, itti2001computational, hu2017squeeze, yu2018learning}. Our AFB is designed to change the weights of different channels in feature maps according to the global context and fuse the adjacent stages information similar to~\cite{hu2017squeeze, yu2018learning}. To achieve this goal, we need the weights which will be used to reweight different channels could respond to different scene context. The previous Encoder-Decoder networks, such as SegNet~\cite{badrinarayanan2017segnet} and U-Net~\cite{ronneberger2015u}, usually ignore the global context and just summed up or concatenate the feature maps of adjacent stages. The convolution operates applied to a feature maps which just summed over all channels can be formed as follows:
    \begin{equation}
        \boldsymbol{y} =\boldsymbol{W} * \boldsymbol{X}  = \sum^{C}_{s=1} \boldsymbol{w}_s * \boldsymbol{x}_s,
    \end{equation}
    where $*$ denotes convolution. $\boldsymbol{y}$ is the output of convolution. $\boldsymbol{X = [x_1, x_2, \ldots, x_c]}$ represents input feature maps and $\boldsymbol{w_s} \in \boldsymbol{W}$ is a 2D spatial kernel which acts on the corresponding channel of $\boldsymbol{X}$. Note that when the training ends, the weights in convolution kernel $\boldsymbol{W}$ are fixed, therefore, they cannot be changed with input in order to give informative features more weight.
    
    As shown in Fig.~\ref{fig:network} (c), the feature maps produced by current Res-Block (represented by dotted line) and previous AFB (represented by solid line) is fused in current AFB. First, the two inputs are concatenated, then feed into a global average pooling layer followed by two $1$x$1$ convolution layers. It produces a $1$x$1$x$C$ tensor in which $C$ is the number of channels in input represented by dotted line. The value in this $1$x$1$x$C$ tensor represents the weight in the corresponding channel of feature maps and it varies according to the context in a different scene. Next, the feature map produced by current Res-Block is weighed by element-wise multiplication with the $1$x$1$x$C$ tensor. Finally, the weighted feature maps is added to the feature maps produced by the previous AFB.
    
    {\bf Channel Reduce Block.} As shown in Fig.~\ref{fig:network} (b), the DFE produces feature maps with huge channels, such as $2048$ in Res-Block $4$. In order to reduce the number of channels and refine the feature maps, we introduce the Channel Reduce Block. The first component of the block is a $1$x$1$ convolution layer in which the input's channel reduced. Then, in order to refine the reduced feature maps, the following is a basic residual block inspired by ResNet~\cite{he2016deep}. In CRB, we use dilated convolution with the same dilation rate as corresponding Res-Block instead of the standard $3$x$3$ convolution.
    
   	\begin{table*}[t]
    \small
    \centering
    \caption{Depth prediction comparison with other methods on NYU Depth V2 dataset.}
    \vspace{1em}
    \begin{tabular}{l|ccc|c|ccc|ccc} 
    \hline
    \multicolumn{4}{l|}{\multirow{2}{*}{Method}}                                                                     & \multirow{2}{*}{Base network} & \multicolumn{3}{c|}{Error (lower is better)}                 & \multicolumn{3}{c}{Accuracy (higher is better)}   \\ 
    \cline{6-6}\cline{7-7}\cline{8-11}
    \multicolumn{4}{l|}{}                                                                                            &                               & rel          & rms           & log10         & $\delta<1.25$ & $\delta<1.25^2$   & $\delta<1.25^3$   \\ 
    \hline
    \multicolumn{4}{l|}{Karsch \etal~\cite{karsch2012depth}}                                                         & -                             & 0.374        & 1.120         & 0.131         & -             & -                 & -      \\
    \multicolumn{4}{l|}{Ladicky \etal~\cite{ladicky2014pulling}}                                                     & -                             & -            & -             & -             & 0.542         & 0.829             & 0.941  \\
    \multicolumn{4}{l|}{Liu \etal~\cite{liu2014discrete}}                                                            & -                             & 0.355        & 1.060         & 0.127         & -             & -                 & -      \\
    \multicolumn{4}{l|}{Li \etal~\cite{li2015depth}}                                                                 & -                             & 0.232        & 0.821         & 0.094         & 0.624         & 0.886             & 0.968  \\
    \multicolumn{4}{l|}{Liu \etal~\cite{liu2015deep}}                                                                & -                             & 0.230        & 0.824         & 0.095         & 0.614         & 0.883             & 0.971  \\
    \multicolumn{4}{l|}{Roy~\cite{roy2016monocular}}                                                                 & -                             & 0.187        & 0.744         & 0.078         & -             & -                 & -      \\
    \multicolumn{4}{l|}{Eigen~\cite{eigen2015predicting}}                                                            & VGG-16                        & 0.158        & 0.641         & -             & 0.769         & 0.950             & 0.988  \\
    \multicolumn{4}{l|}{Chakrabarti \etal~\cite{chakrabarti2016depth}}                                               & VGG-19                        & 0.149        & 0.620         & -             & 0.806         & 0.958             & 0.987  \\
    \multicolumn{4}{l|}{Laina \etal~\cite{laina2016deeper}}                                                 & ResNet-50                     & \third{0.127}& \third{0.573} & 0.055         & 0.811         & 0.953             & 0.988  \\
    \multicolumn{4}{l|}{Li \etal~\cite{li2017two}}                                                                   & VGG-16                        & 0.152        & 0.611         & 0.064         & 0.789         & 0.955             & 0.988  \\
    \multicolumn{4}{l|}{MS-CRF~\cite{xu2017multi}}                                                                   & ResNet-50                     & \best{0.121} & 0.586         & \best{0.052}  & 0.811         & 0.954             & 0.987  \\
    \multicolumn{4}{l|}{Xu \etal~\cite{xu2018structured}}                                                            & ResNet-50                     &\second{0.125}& 0.593         & 0.057         & 0.806         & 0.952             & 0.986  \\ 
    \hline
    \multirow{5}{*}{Ours}                        & \multicolumn{1}{l|}{DFE} & \multicolumn{1}{l|}{AFB} & Dilated CRB &                               &              &               &               &               &                   &        \\ 
    \cline{2-11}
                                                 &                          &                          &             & ResNet-50                     & 0.153        & 0.663         & 0.070         & 0.754         & 0.953             & \second{0.990}  \\
                                                 & \checkmark               &                          &             & ResNet-101                    & 0.134        & 0.583         & 0.056         & \third{0.826} & \third{0.962}     & \third{0.989}  \\
                                                 & \checkmark               & \checkmark               &             & ResNet-101                    & 0.129        & \second{0.568}& \third{0.054} & \second{0.836}& \second{0.965}    & \best{0.991}  \\
                                                 & \checkmark               & \checkmark               & \checkmark  & ResNet-101                    & \third{0.127}& \best{0.555}  & \second{0.053}& \best{0.841}  & \best{0.966}      & \best{0.991}  \\
    \hline
    \end{tabular}
    \label{table:compare}
    \end{table*}
    
    \subsection{Network Architecture}

    By combining DFE and DMG, we propose our complete network for predicting depth from a single image as illustrated in Fig.~\ref{fig:network} (a).

    We use ResNet~\cite{he2016deep} pre-trained on ImageNet~\cite{deng2009imagenet} as a backbone of our DFE. As shown in the diagram, there are four different Res-Block in our DFE according to the dilation rate $r \in \{1, 2, 4, 8\}$ respectively. The final feature maps in first three Res-Block is feed into a CRB, then, an AFB followed by another CRB. Due to lacking followed Res-Block, the last Res-Block is feed into only a CRB. The feature maps in all Res-Block have the same spatial resolution ($1/4$ of the input image) because we use dilated convolution instead of standard strided convolution. Therefore, we apply only one $2$x bilinearly upsample to the last CRB output in order to get a denser depth map.

    In training, we use $\mathcal{L}_1$ Loss in $log$ space defined between ground truth depth map $D$ and predicted depth map $\hat{D}$ as:
    \begin{equation}
        L(D, \hat{D}) = \frac{1}{n} \sum^n_p\lvert \log_e(D_p+1) - \log_e(\hat{D}_p+1)\rvert ,
    \end{equation}
where $n$ is the number of pixels in depth map and $p$ is a pixel index. Converting depth to $log$ space can down-weight contribution of regions with large depth value. It benefits for training because the regions with larger depth value have less rich information for depth estimation.
    
    \subsection{Implementation Details}
    We use TensorFlow~\cite{abadi2016tensorflow} framework to implement the proposed model and train on a single NVIDIA TITAN Xp GPU with 12GB memory. We initialize the DFE of our proposed network with ResNet-101 weights which pre-trained for ILSVRC image classification task and the weights in DMG are randomly initialized with Xavier initializer~\cite{glorot2010understanding}. After initialization, we fixed all parameters in the first two stages in ResNet-101 because the first few layers extract general low-level information, and it could be beneficial to prevent overfitting. Besides, the batch normalization parameters are fixed during training time. The batch size is set to 3. Training is performed for $20$ epochs, which consists of $50K$ steps per epoch. Similar to \cite{chen2018encoder}, we use a ``poly" learning rate decay policy with Adam~\cite{kinga2015method} optimizer in which the learning rate $l$ is defined as:
    \begin{equation}
        l = (l_{init} - l_{end}) * (1 - \frac{global\_step}{decay\_steps})^{power} + l_{end} ,
    \end{equation}
    where $l_{init}$ is initial learning rate and $l_{end}$ is end learning rate, $global\_step = \min(global\_step, decay\_steps)$. Specifically, we set $l_{init} = 10^{-5}$ and $l_{end} = 10^{-7}$ in DFE, $l_{init} = 10^{-4}$ and $l_{end} = 10^{-6}$ in DMG. Since the parameters in DFE are pre-trained on ImageNet, they need more fine scale change. The $decay\_steps$ is set to $16$ epochs and $power$ is set to $1.0$ in both DFE and DMG. Our training time is about $50$ hours in our sampled training set.

\section{Experiments}

\subsection{Dataset and Evaluation Metrics}

    {\bf Dataset.} We use one of the largest indoor scene dataset, NYU Depth V2, for training and evaluation. The NYU Depth V2 dataset contains two sub-datasets: labeled dataset and raw dataset. The raw dataset consists of $249$ training scenes and $215$ test scenes captured with Microsoft Kinect. The labeled dataset has $1449$ aligned RGB-D pairs which be officially split into two parts: $795$ pairs for training and $654$ for testing. For training our depth estimation network, we extract ${\sim}13K$ pairs from $249$ training scenes in raw dataset. The images in the dataset which have the resolution in $480\times640$ are down-sampled to $312 \times 416$ and then cropped to $240 \times 320$ in order to remove blank boundaries. We also perform randomly online data augmentation for the training pairs as follows.
    \begin{enumerate}
    	\item The images RGB value are added with a random \hbox{$\Delta \in [-0.2, 0.2]$} to adjust the image brightness.
        \item The images contrast are adjusted with a factor \hbox{$\delta \in [0.8, 1.2]$}.
        \item Both RGB images and depth maps are horizontally flipped with $0.5$ probability.
    \end{enumerate}
    
    Similar to previous works~\cite{eigen2015predicting, li2015depth, liu2014discrete, xu2017multi}, we evaluate our predicted depth map using mean relative error (rel), root mean squared error (rms), mean $\log_{10}$ error and accuracy with threshold as evaluation metrics.
	
    \begin{figure*}[t]
        \centering
            \includegraphics[width=\linewidth]{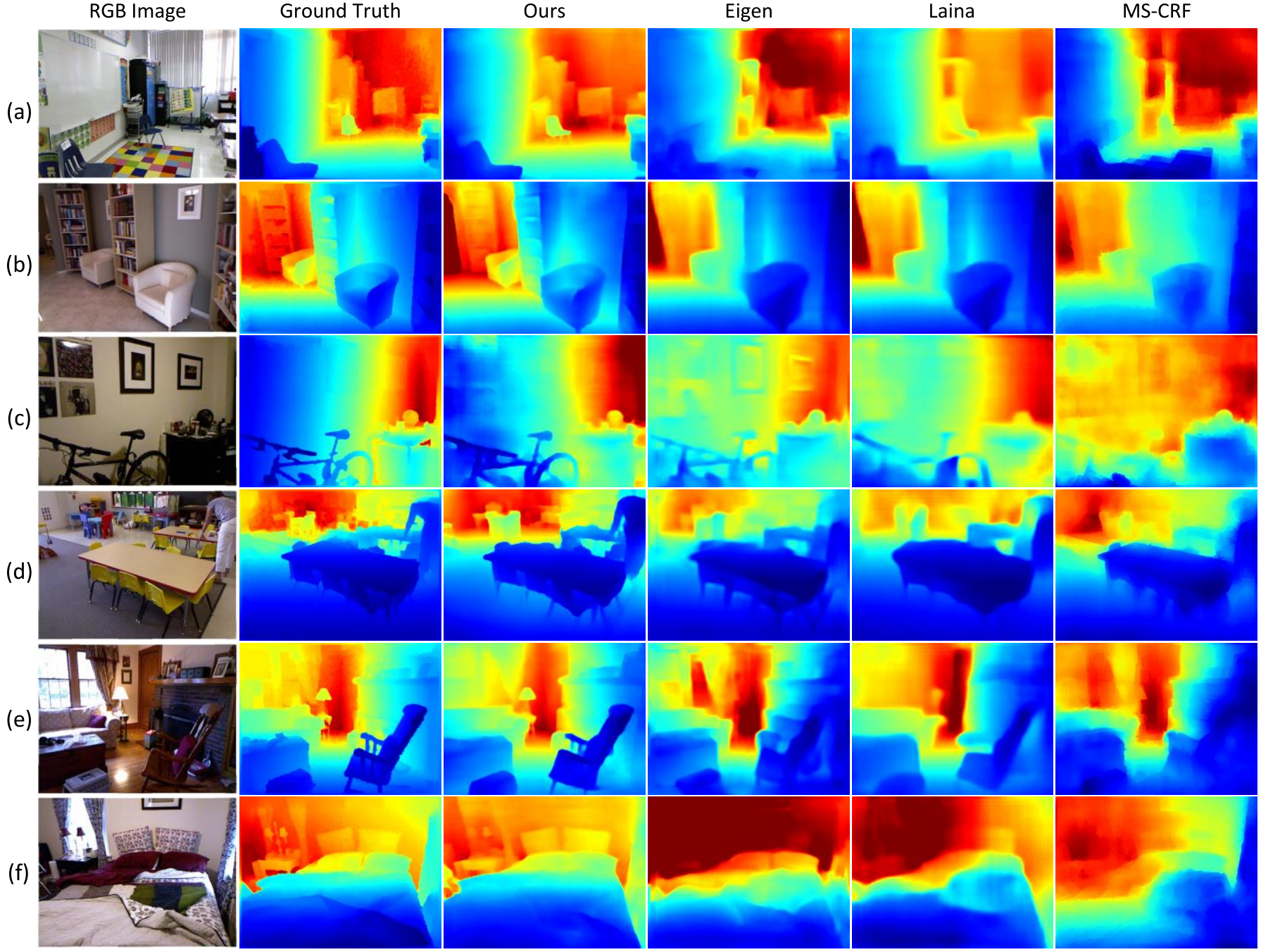}
        \caption{{\bf Qualitative Comparisons on NYU Depth.} Our predictions are clearer and sharper than Eigen~\cite{eigen2015predicting},
        Laina~\cite{laina2016deeper} and MS-CRF~\cite{xu2017multi}.}
        \label{fig:visual}
    \end{figure*}
    \begin{figure}[tb]
        \centering
            \includegraphics[width=\linewidth]{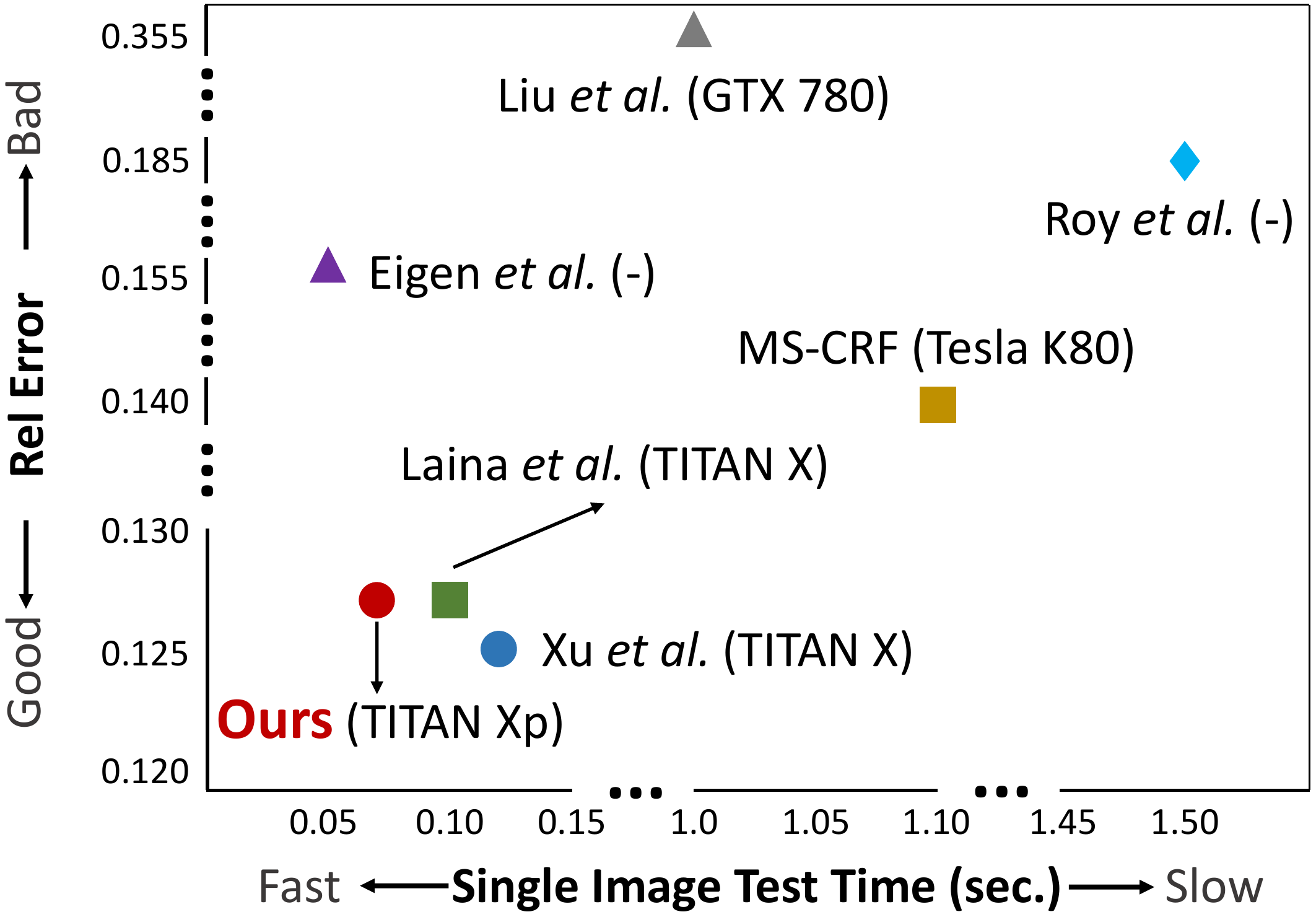}
        \caption{{\bf Accuracy and Speed Comparison.} Comparison with previous method: rel error versus a single image prediction time. Numbers in char are taken from \cite{xu2018structured}.}
        \label{fig:test_time}
    \end{figure}
    
\subsection{Performance on NYU Depth V2}
    {\bf Comparison with the State-of-the-art.} In Table~\ref{table:compare} we compare the accuracy of depth predicted by the proposed network with the state-of-the-art. The numerical results are taken from the corresponding original paper. The best, second and third result on each metric is colored in \textcolor{tableRed}{red}, \textcolor{tableGre}{green} and \textcolor{tableBlu}{blue} respectively. Our method outperforms all other methods in accuracy metrics. In error metrics, we obtain the best rms and competitive rel and log10 results. We would like to point out that MS-CRF~\cite{xu2017multi} which achieve the best rel and log10 error is trained on a $95K$ subset of the NYU Depth V2 dataset. While our train set, consists of $13K$ samples, is radically smaller than MS-CRF.
    
    {\bf Ablation Study.} We use a simple encoder-decoder architecture consists of an original ResNet-50 removed the global pooling and fully connected layer in the end as encoder and stacked deconvolution layers as decoder as the baseline. In the baseline network, we also introduce skip connection~\cite{ronneberger2015u} between the same spatial resolution feature maps in encoder and in decoder. The performance of the baseline network is reported in the first row of our methods in Table~\ref{table:compare}. Comparing the baseline with the proposed network, it clear that leveraging DFE and DMG improves the performance by a large margin. To evaluate the effect of our proposed modules, we add DFN, AFB and dilated CFB to the baseline network gradually. The results show that each of our proposed modular can improve the performance of the corresponding network.

    {\bf Qualitative Comparison.} In Fig.~\ref{fig:visual} we show some selected visual results of our best model and compare with the publicly available predictions of Eigen~\cite{eigen2015predicting}, Laina \etal~\cite{laina2016deeper} and MS-CRF~\cite{xu2017multi}. One can clearly see that depth maps produced by our method have noteworthy visual quality especially edge quality and rich details, for example, in the $(c)$ row, the bike in our depth map is as clear as in ground truth. Moreover, in the $(d)$ row, the person in our depth map is sharp, while, cannot be able to tell in other methods. In the $(a)$ row, our depth map is the only one which the chair in far is able to tell from the background. It is also worth noting that our method produces depth map end-to-end, thus, do not need any additional post-processing such as CRF.

    {\bf Efficiency.} Robotics is one of the major application areas that indoor scene depth estimation can be used for. In which prediction speed is critical. Since we use DFE, there is only one $2$x upsampling in our DMG which allow our framework to work at ${\sim}15fps$ for inferencing. We compare the proposed method with previous methods in terms of computational cost in test phase in Fig.~\ref{fig:test_time}. Our approach outperforms Roy \etal~\cite{roy2016monocular}, Liu \etal~\cite{liu2014discrete} and MS-CRF (small training set)~\cite{xu2017multi} both in terms of accuracy and of running time. Furthermore, in comparison to Eigen \etal~\cite{eigen2015predicting}, Laina \etal~\cite{laina2016deeper} and Xu \etal~\cite{xu2018structured}, the proposed framework obtains an impressive trade-off between accuracy and speed.
    
    \begin{figure*}[tb]
        \centering
            \includegraphics[width=0.9\linewidth]{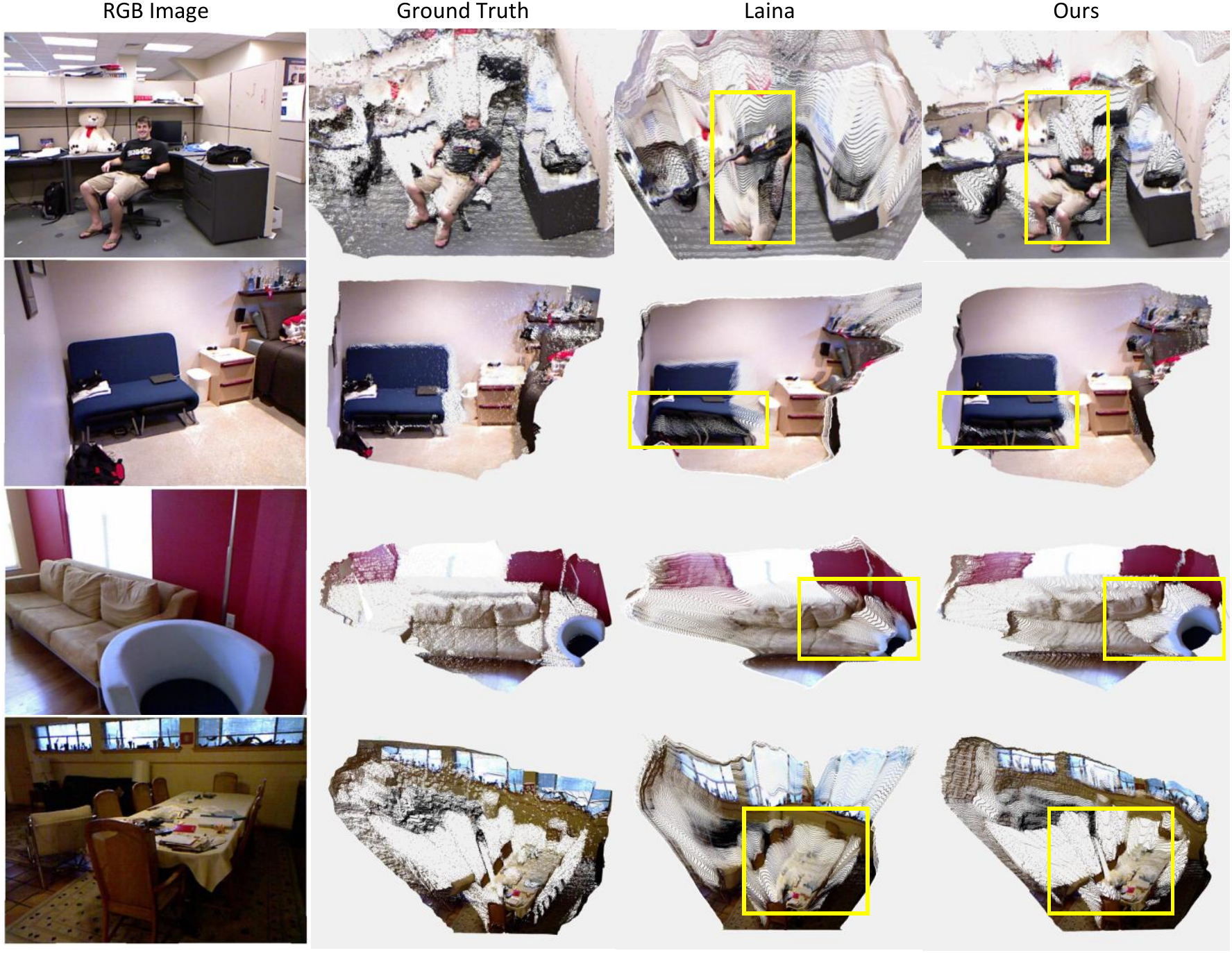}
        \caption{{\bf Projecting Depth Map to 3D Point Cloud Comparisons.} Our results recover better 3D information.}
        \label{fig:point_cloud}
    \end{figure*}

\subsection{Applications: point cloud and bokeh}
    To further illustrate the differences between our method and the previous methods, and emphasize the importance of sharp
    edge, we compare our result with Laina \etal~\cite{laina2016deeper} in (1) 3D point cloud, and (2) application of
    bokeh.
    
    \begin{figure}[tb]
        \centering
            \includegraphics[width= 0.9\linewidth]{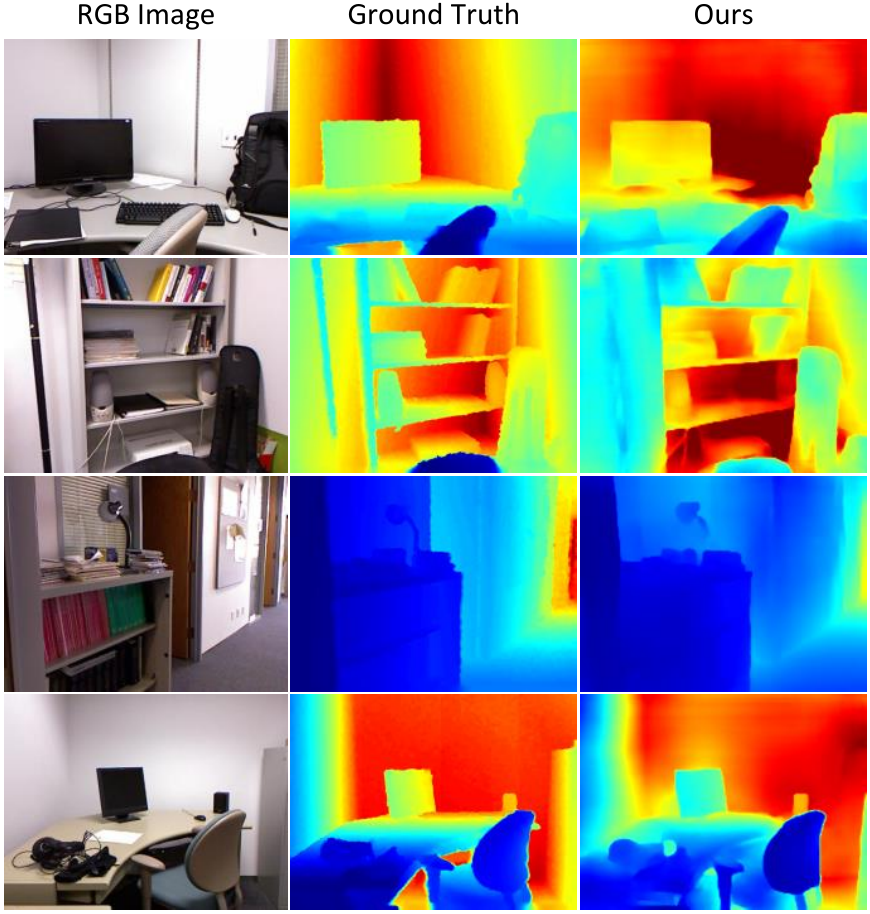}
        \caption{Representative results on SUN-RGBD dataset.}
        \label{fig:sunrgbd}
    \end{figure}
    
    \begin{figure*}[tb]
        \centering
            \includegraphics[width=0.8\linewidth]{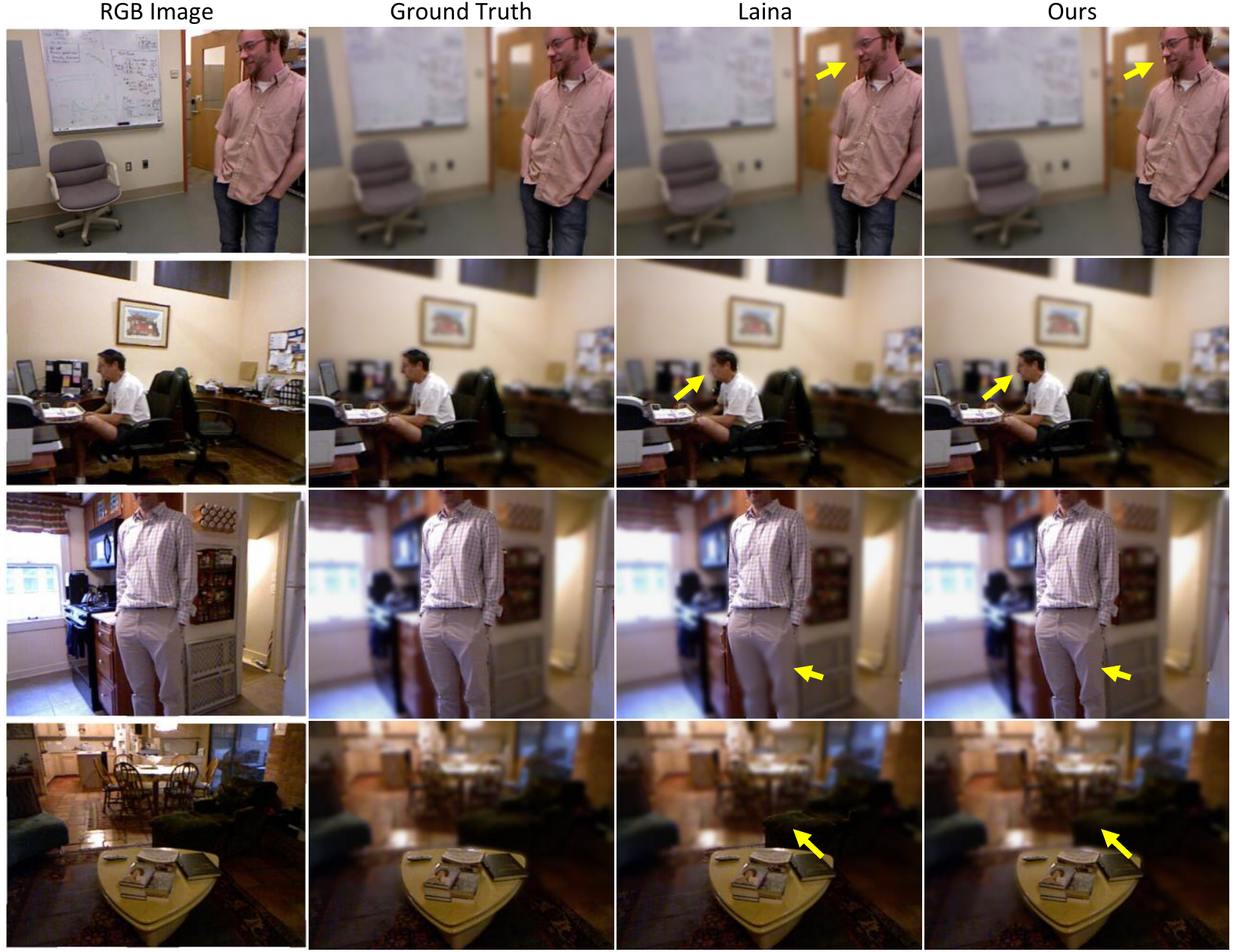}
        \caption{{\bf Bokeh Generation Comparisons.} Depth maps estimated by our method produce better blur effect.}
        \label{fig:bokeh}
    \end{figure*}

{\bf Point Cloud.} We make a qualitative comparison in $3D$ by projecting the predicted $2D$ depth map into $3D$ point cloud. The coordinate of the point is computed by a script provided by NYU Depth V2 toolbox. To visualize the quality of produced point cloud, we render the point cloud in a different view from the original RGB image. Some selected results are shown in Fig.~\ref{fig:point_cloud}. It is easy to see that our depth map produces a better $3D$ representation of the scene. For example, in the first row, our method predicts the person in a relative right place while Laina \etal~\cite{laina2016deeper} predict the head far away from the body. It also worth noting that Laina reports a similar evaluate score to ours, one can see that a numerical metrics such as rel not always reveal the predicted depth quality.

{\bf Bokeh.} Bokeh effect is usually obtained with DSLR camera and requires special photography skill. It is a highly desired feature in consumer level like photography using a smart phone. This is now feasible with the scene depth information. Some high-end phones equip with additional sensors for capturing the depth (e.g. iPhone's portrait mode). We show that based on our depth prediction method we can render realistic bokeh effect just with a single color image. With the estimated depth map, we can choose to either keep sharp contents (`in-focus region') or blur the scenes (`out-of-focus region') according to their depth planes. Fig.~\ref{fig:bokeh} shows four examples. It is obvious that the results using our depth prediction method is closer to the ones using ground truth depth which keep the edge of object sharper. Specifically, in first to third rows, our depth map makes the person stand out from the background while Laina \etal~\cite{laina2016deeper} fuses partial body with the background (legs in third rows, face in the first and second row).
    
\subsection{Generalization}
    
    To show the generalization of the proposed network, we test it on B3DO~\cite{janoch2013category} which is contained in SUN-RGBD dataset~\cite{song2015sun}. The image and depth in B3DO are captured by a Microsoft Kinect which has similar camera parameters to the camera used in NYU Depth V2. Note that our depth estimate network is trained on data only contains samples in NYU Depth V2 and never see images in B3DO. Fig.~\ref{fig:sunrgbd} shows some result samples that our network predicts in B3DO. One can see that our network obtains a quite good result especially sharp edge structures.

\section{Conclusions}
	
    In this paper, we have proposed a novel approach for single image depth estimation. Our method is motivated by obtaining a detailed depth map. Unlike previous works that typically reduce spatial resolution of the feature maps, our method introduces DFE to extract features from the image while keeping the feature maps dense. Moreover, attention mechanism is integrated into DMG to fuse the multi-scale information produced in DFE. We evaluate the proposed method both numerically and qualitatively on the benchmark dataset. By performing applications such as bokeh generation and $3D$ point cloud, it indicates that our result has more fine-scaled details and align with depth boundary better. In the future, we want to extend our method to estimate depth in monocular video with temporal consistency.
    
\section*{Acknowledgements}
	
     This work was supported by NSFC under Grant 61602020 and Grant 61732016. The GPU is granted by DRIVE PX Grant Program.
	

{\small
\bibliographystyle{ieee}
\bibliography{Detail_Preserving_Depth_Estimation_from_a_Single_Image_using_Attention_Guided_Networks}
}

\end{document}